\documentclass[11pt,a4paper]{article}
\usepackage[printwatermark]{xwatermark}
\usepackage[hyperref]{emnlp-ijcnlp-2019}
\usepackage{times}
\usepackage{latexsym}
\usepackage{xcolor}
\usepackage{url}
\usepackage{enumitem}
\usepackage{graphicx}
\usepackage{amssymb}
\usepackage{booktabs}
\usepackage{multirow,multicol}
\usepackage{xspace}

\aclfinalcopy 
\newcommand{\papername}{\textsc{Social\hspace{.15em}IQa}\xspace}
\newcommand{\fulldataname}{Social Intelligence QA}

\newcommand\aitwo{$^\diamondsuit$}
\newcommand\uw{$^\heartsuit$}
\newcommand\coauth{$^\star$~}
\newcommand\aspace{\hspace{.75em}}
\title{
    \papername: 
    Commonsense Reasoning about Social Interactions
}

\author{
 Maarten Sap\coauth \aitwo\uw \aspace
 Hannah Rashkin\coauth \aitwo\uw \aspace
 Derek Chen\uw \aspace
 Ronan Le Bras\aitwo \aspace
 Yejin Choi\aitwo\uw \\
 \aitwo Allen Institute for Artificial Intelligence, Seattle, WA, USA\\
 \uw Paul G. Allen School of Computer Science \& Engineering, Seattle, WA, USA\\
  {\tt \{msap,hrashkin,dchen14,yejin\}@cs.washington.edu} \\
  {\tt \{ronanlb\}@allenai.org}
}

\begin{document}

\maketitle
\begin{abstract}
    We introduce \papername, the first large-scale benchmark for commonsense reasoning about social situations.
    \papername contains 38,000 multiple choice questions for probing \textit{emotional} and \textit{social} intelligence in a variety of everyday situations (e.g., 
    Q:~``\textit{Jordan wanted to tell Tracy a secret, so Jordan leaned towards Tracy. Why did Jordan do this?}'' A:~``\textit{Make sure no one else could hear}'').
    Through crowdsourcing, we collect commonsense questions along with correct and incorrect answers about social interactions, using a new framework that mitigates stylistic artifacts in incorrect answers by asking workers to provide the right answer to a different but related question.
    Empirical results show that our benchmark is challenging for existing question-answering models based on pretrained language models, compared to human performance ($>$20\% gap).
    Notably, we further establish \papername as a resource for transfer learning of commonsense knowledge,  
    achieving state-of-the-art performance on multiple commonsense reasoning tasks (Winograd Schemas, COPA).
    
    \let\thefootnote\relax\footnotetext{\coauth Both authors contributed equally.}
\end{abstract}

\section{Introduction}
\begin{figure}[t]
    \centering
    \includegraphics[width=\columnwidth]{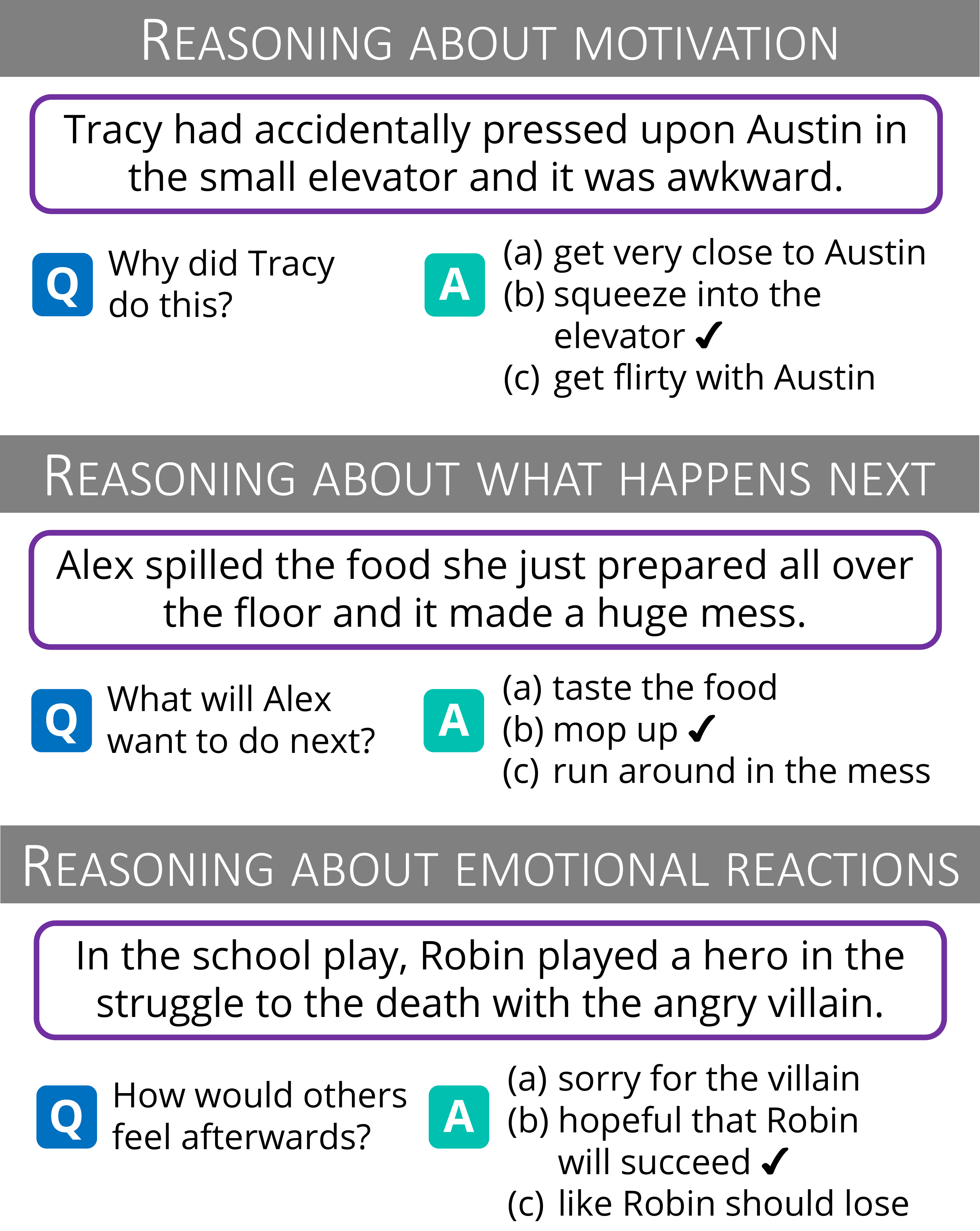}
    \caption{
    Three context-question-answers triples from \papername, along with the type of reasoning required to answer them.
    In the top example, humans can trivially infer that Tracy pressed upon Austin because there was no room in the elevator.
    Similarly, in the bottom example, commonsense tells us that people typically root for the hero, not the villain.
    }
    \label{fig:intro-fig}
\end{figure}

Social and emotional intelligence enables humans to reason about the mental states of others and their likely actions~\cite{ganaie2015study}. 
For example, when someone spills food all over the floor, we can infer that they will likely want to 
clean up the mess, rather than taste the food off the floor or run around in the mess (Figure~\ref{fig:intro-fig}, middle). 
This example illustrates how Theory of Mind, i.e., the ability to reason about the implied emotions and behavior of others, enables humans to navigate social situations ranging from simple conversations with friends to complex negotiations in courtrooms~\cite{apperly2010mindreaders}.


While humans trivially acquire and develop such social reasoning skills~\cite{moore2013development}, this is still a challenge for machine learning models, 
in part due to the lack of large-scale resources to train and evaluate modern AI systems' social and emotional intelligence.
Although recent advances in pretraining large language models have yielded promising improvements on several commonsense inference tasks, these models still struggle to reason about social situations, as shown in this and previous work \cite{Davis2015CommonsenseRA,Nematzadeh2018EvaluatingToM,Talmor2019commonsenseQA}.
This is partly due to language models being
trained on written text corpora, where reporting bias of knowledge limits the scope of commonsense knowledge that can be learned \cite{Gordon2013reportingbias,Lucy2017AreDR}.


In this work, we introduce \fulldataname{} (\papername), the first large-scale resource to learn and measure \textit{social} and \textit{emotional} intelligence in computational models.\footnote{Available at \url{https://tinyurl.com/socialiqa}}
\papername contains 38k multiple choice questions
regarding the pragmatic implications of everyday, social events (see Figure~\ref{fig:intro-fig}).  To collect this data, we design a crowdsourcing framework to gather contexts and questions that explicitly address social commonsense reasoning.
Additionally, by combining handwritten negative answers with adversarial question-switched answers (Section~\ref{sec:qsa}), we minimize annotation artifacts that can arise from crowdsourcing incorrect answers~\cite{Schwartz2017effect,Gururangan2018AnnotationAI}. 


This dataset remains challenging for AI systems, with our best performing baseline reaching 64.5\% (BERT-large), significantly lower than human performance.
We further establish \papername as a resource that enables transfer learning for other commonsense challenges, through \textit{sequential finetuning} of a pretrained language model on \papername before other tasks.
Specifically, we use \papername to set a new state-of-the-art on three commonsense challenge datasets: COPA \cite{Roemmele2011ChoiceOP} (83.4\%), the original Winograd \citep{Levesque2011TheWS} (72.5\%), and the extended Winograd dataset from \citet{Rahman2012DPR} (84.0\%). 

Our contributions are as follows: (1) We create \papername,
the first large-scale QA dataset aimed at testing social and emotional intelligence, containing over 38k QA pairs.
(2) We introduce question-switching, a technique to collect incorrect answers that minimizes stylistic artifacts due to annotator cognitive biases.
(3) We establish baseline performance on our dataset, with BERT-large performing at 64.5\%, well below human performance.
(4) We achieve new state-of-the-art accuracies on COPA and Winograd through sequential finetuning on \papername, which 
implicitly endows models with social commonsense knowledge.

\begin{table}[t]
    \centering
    \begin{tabular}{@{}llr@{}}
    \toprule
\multicolumn{3}{@{}c@{}}{\papername} \\ \midrule
    \multirow{4}{2.3cm}{\# QA tuples} & train & 33,410 \\ 
    & dev & 1,954 \\
    & test & 2,224  \\
    & \textbf{total} & \textbf{37,588} \\\bottomrule
    \toprule
    \multicolumn{3}{@{}c@{}}{Train statistics} \\ \midrule
    \multirow{5}{2.3cm}{Average \\\# tokens} & context & 14.04 \\
        & question & 6.12  \\
        & answers (all) & 3.60 \\
        & answers (correct) & 3.65 \\
        & answers (incorrect) & 3.58  \\ \midrule
    \multirow{4}{2.3cm}{Unique \\\# tokens} & context & 15,764\\
        & question &  1,165 \\
        & answers (all)&  12,285\\
        & answers (correct) & 7,386\\
        & answers (incorrect) &  10,514\\ \midrule
        \multirow{2}{2.3cm}{Average freq. of answers} & answers (correct) & 1.37\\
        & answers (incorrect) & 1.47 \\ 
    \bottomrule
    \end{tabular}
    \caption{Data statistics for \papername.}
    \label{tab:data-stats}
\end{table}
\section{Task description}
\papername aims to measure the social and emotional intelligence of computational models through multiple choice question answering (QA).
In our setup, models are confronted with a question explicitly pertaining to an observed \textit{context}, where the correct answer can be found among three competing options. 

By design, the questions 
require \textit{inferential} reasoning about the social causes and effects of situations, 
in line with the type of intelligence required for an AI assistant to interact with human users \citep[e.g., know to call for help when an elderly person falls; ][]{Pollack2005IntelligentTF}. 
As seen in Figure~\ref{fig:intro-fig}, correctly answering questions requires reasoning about motivations, emotional reactions, or likely preceding and following actions.
Performing these inferences is what makes us experts at navigating social situations, and is closely related to Theory of Mind, i.e., the ability to reason about the beliefs, motivations, and needs of others \cite{baron1985SallyAnne}.\footnote{
Theory of Mind is well developed in most neurotypical adults \cite{ganaie2015study}, but can be influenced by 
age, culture, or developmental disorders \cite{korkmaz2011theory}.}
Endowing machines with this type of intelligence has been a longstanding but elusive goal of AI \cite{Gunning2018machineCommonsense}. 



\subsection*{ATOMIC}
As a starting point for our task creation, we draw upon social commonsense knowledge from ATOMIC \cite{sap2019atomic} to seed our contexts and question types.
ATOMIC is a large knowledge graph that contains inferential knowledge about the causes and effects of 24k short events.
Each triple in ATOMIC consists of an event phrase with person-centric variables, one of nine inference dimensions, and an inference object (e.g., ``PersonX pays for PersonY's \_\_\_'', ``xAttrib'', ``generous'').
The nine inference dimensions in ATOMIC cover causes of an event (e.g., ``X needs money''), its effects on the agent (e.g., ``X will get thanked'') and its effect on other participants (e.g., ``Y will want to see X again''); see \citet{sap2019atomic} for details.

Given this base, we generate natural language contexts that represent specific instantiations of the event phrases found in the knowledge graph.  Furthermore, the questions created probe the commonsense reasoning required to navigate such contexts.  Critically, since these contexts are based off of ATOMIC, they explore a diverse range of motivations and reactions, as well as likely preceding or following actions.



\section{Dataset creation}
\papername contains 37,588 multiple choice questions with three answer choices per question.
Questions and answers are gathered through three phases of crowdsourcing aimed to collect the \textit{context}, the \textit{question}, and a set of \textit{positive} and \textit{negative} answers.
We run crowdsourcing tasks on Amazon Mechanical Turk (MTurk) to create each of the three components, as described below.

\subsection{Event Rewriting}
In order to cover a variety of social situations, we use the base events from ATOMIC as prompts for context creation. 
As a pre-processing step, we run an MTurk task that asks workers to turn an ATOMIC event (e.g., ``PersonX spills \_\_\_ all over the floor'') into a sentence by adding names, fixing potential grammar errors, and filling in placeholders (e.g., ``Alex spilled food all over the floor.'').
\footnote{This task paid \$0.35 per event.}

\begin{figure}[t]
    \centering
    \includegraphics[width=\columnwidth]{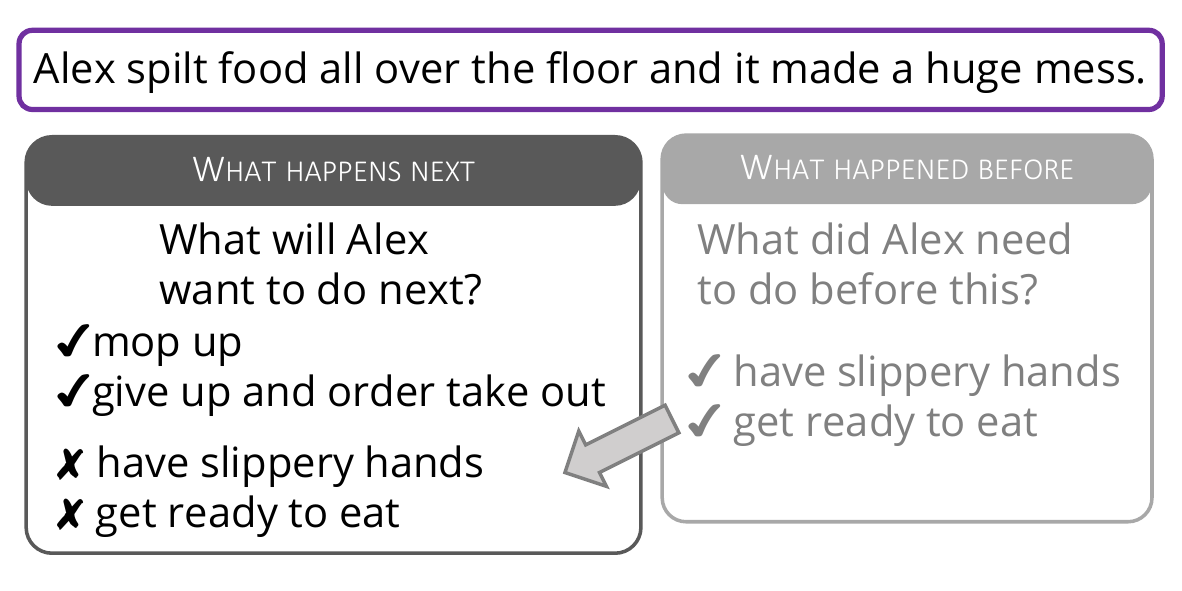}
    \caption{Question-Switching Answers (QSA) are collected as the correct answers to the wrong question that targets a different type of inference (here, reasoning about what happens before instead of after an event).
    }
    \label{fig:question-switching}
\end{figure}

\subsection{Context, Question, \& Answer Creation}
Next, we run a task where annotators create full context-question-answers triples.
We automatically generate question templates covering the nine commonsense inference dimensions in ATOMIC.\footnote{We do not generate templates if the ATOMIC dimension is annotated as ``none.''}
Crowdsourcers are prompted with an event sentence and an inference question to turn into a more detailed context\footnote{Workers were asked to contribute a context 7-25 words longer than the event sentence.} (e.g. ``Alex spilled food all over the floor and it made a huge mess.'') and an edited version 
of the question if needed for improved specificity
(e.g. ``What will Alex want to do next?'').
Workers are also asked to contribute two potential correct answers. 


\begin{figure*}[t]
    \centering
    \includegraphics[width=\textwidth]{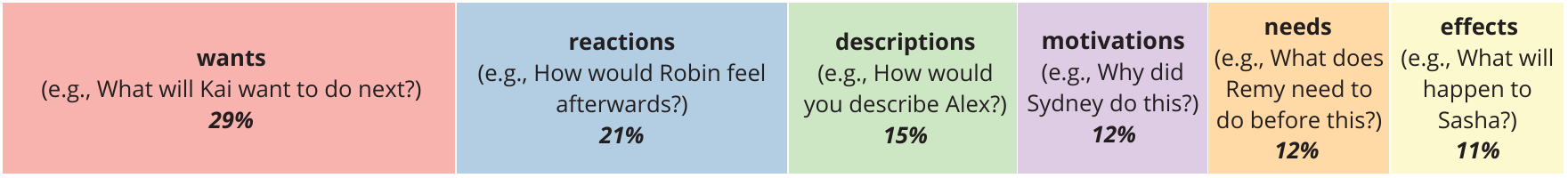}
    \caption{\papername contains several question types which cover different types of inferential reasoning. Question types are derived from ATOMIC inference dimensions.}
    \label{fig:dimension-breakdown}
\end{figure*}

\subsection{Negative Answers}
In addition to correct answers, we collect four incorrect answer options, of which we filter out two.
To create incorrect options that are adversarial for models but easy for humans, we use two different approaches to the collection process.
These two methods are 
specifically designed to avoid different types of annotation artifacts, thus making it more difficult for models to rely on data biases.
We integrate and filter answer options and validate final QA tuples with human rating tasks.

\paragraph{Handwritten Incorrect Answers (HIA)}
The first method involves eliciting handwritten incorrect answers that require reasoning about the context.  These answers are handwritten to be similar to the correct answers in terms of topic, length, and style but are subtly incorrect.  Two of these answers are collected during the same MTurk task as the original context, questions, and correct answers. We will refer to these negative responses as handwritten incorrect answers (HIA).

\paragraph{Question-Switching Answers (QSA)}
\label{sec:qsa}
We collect a second set of negative (incorrect) answer candidates by switching the questions asked about the context, as shown in Figure~\ref{fig:question-switching}.
We do this to avoid cognitive biases and annotation artifacts in the answer candidates, such as those caused by writing incorrect answers or negations~\cite{Schwartz2017effect,Gururangan2018AnnotationAI}.
In this crowdsourcing task, we provide the same context as the original question, as well as a question automatically generated from a different but similar ATOMIC dimension,\footnote{Using the following three groupings of ATOMIC dimensions: \{xWant, oWant, xNeed, xIntent\}, \{xReact oReact, xAttr\}, and \{xEffect, oEffect\}.} and ask workers to write two correct answers.  We refer to these negative responses as question-switching answers (QSA).

\begin{figure}
    \centering
    \includegraphics[width=\columnwidth]{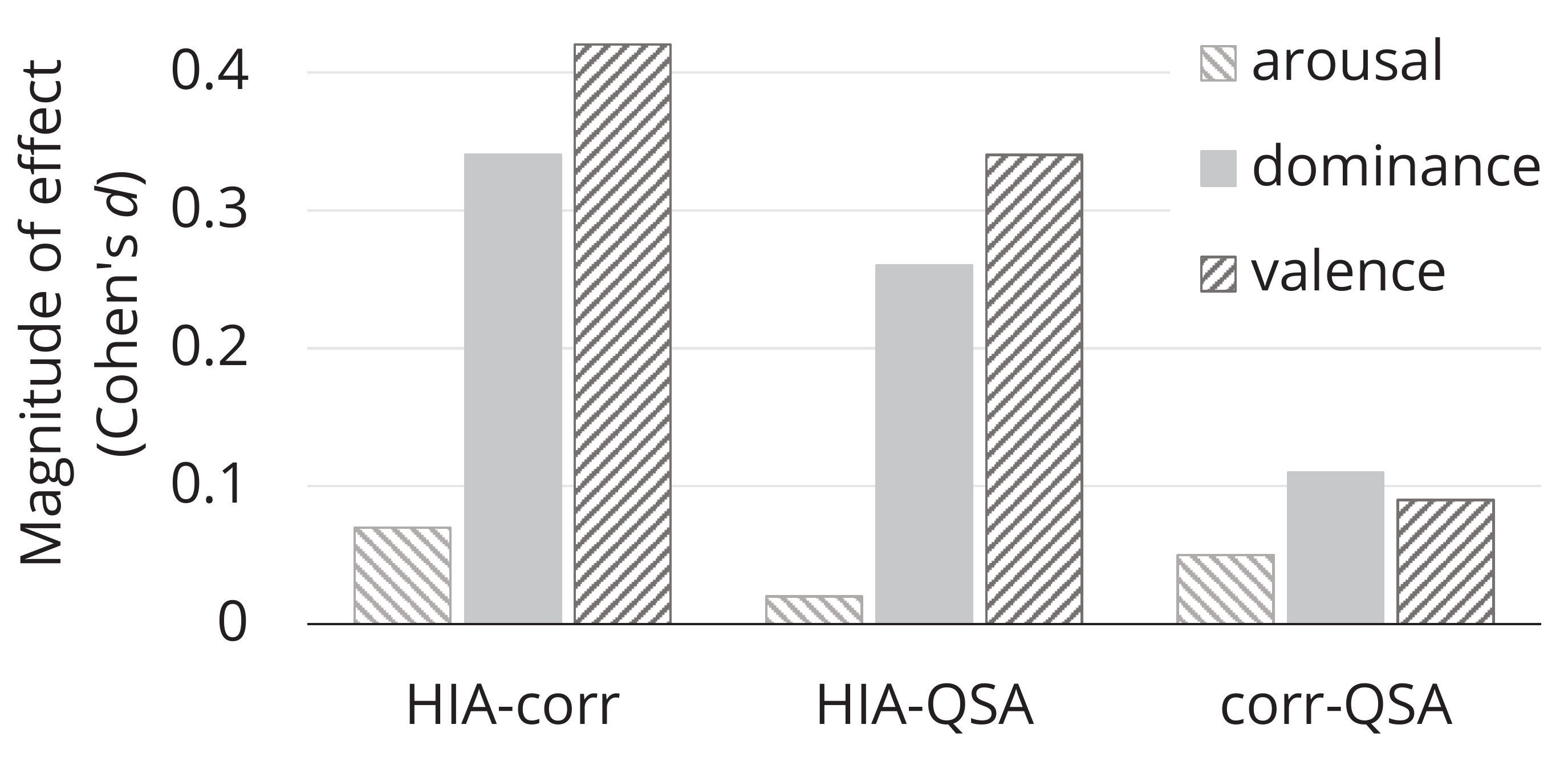}
    \caption{
    Magnitude of effect sizes (Cohen's $d$)
    when comparing average dominance, arousal and valence values of different answer types where larger $|d|$ indicates more stylistic difference.
    For valence (sentiment polarity) and dominance, the effect sizes comparing QSA and correct answers are much smaller, indicating that these are more similar tonally.
    Notably, all three answer types have comparable levels of arousal (intensity).
    }
    \label{fig:answer-sentiment-effects}
\end{figure}

By including answers to a different question about the same context, we ensure that these adversarial responses have the stylistic qualities of correct answers and strongly relate to the context topic, while still being incorrect, making it difficult for models to simply perform pattern-matching.
To verify this, we compare valence, arousal, and dominance (VAD) levels across answer types, computed using the VAD lexicon by \citet[][]{mohammad2018NRC-VAD}.
Figure \ref{fig:answer-sentiment-effects} shows effect sizes (Cohen's $d$) of the differences in VAD means, where the magnitude of effect size indicates how different the answer types are stylistically.
Indeed, QSA and correct answers differ substantially less than HIA answers ($|d|{\leq}.1$).\footnote{Cohen's $|d|{<}.20$ is considered small \cite{Sawilowsky2009NewES}.
We find similarly small effect sizes using other sentiment/emotion lexicons.}

\subsection{QA Tuple Creation}
As the final step of the pipeline, we aggregate the data into three-way multiple choice questions.
For each created context-question pair contributed by crowdsourced workers, we select a random correct answer and the incorrect answers that are least entailed by the correct one, following inspiration from \citet{Zellers2019vcr}.

For the training data, we validate our QA tuples through a multiple-choice crowdsourcing task where three workers are asked to select the right answer to the question provided.\footnote{Agreement on this task was high (Cohen's $\kappa$=.70)}
In order to ensure even higher quality, we validate the dev and test data a second time with five workers.
Our final dataset contains questions for which the correct answer was determined by human majority voting, discarding cases without a majority vote.
We also apply a lightweight form of adversarial filtering to make the task more challenging  by using a deep stylistic classifier to remove easier examples on the dev and test sets \citep[][]{Winogrande}.\footnote{We also tried filtering to remove examples from the training set but found it did not significantly change performance.
We will release tags for the easier training examples with the full data.}

To obtain human performance, we run a separate task asking three new workers to select the correct answer on a random subset of 900 dev and 900 test examples.
Human performance on these subsets is 87\% and 84\%, respectively.


\subsection{Data Statistics}
To keep contexts separate across train/dev/test sets, we assign \papername contexts to the same partition as the ATOMIC event the context was based on.
Shown in Table~\ref{tab:data-stats} (top), this yields a total set of around 33k training, 2k dev, and 2k test tuples.
We additionally include statistics on word counts and vocabulary of the training data.
We report the averages of correct and incorrect answers in terms of: token length, number of unique tokens, and number of times a unique answer appears in the dataset.
Note that due to our three-way multiple choice setup, there are twice as many incorrect answers which influences these statistics.

We also include a breakdown (Figure~\ref{fig:dimension-breakdown}) across question types, which we derive from ATOMIC inference dimensions.\footnote{We group agent and theme ATOMIC dimensions together (e.g., ``xReact'' and ``oReact'' become the ``reactions'' question type).}
In general, questions relating to what someone will feel afterwards or what they will likely do next are more common in \papername. 
Conversely, questions pertaining to (potentially involuntary) effects of situations on people are less frequent.


\section{Methods}
We establish baseline performance on \papername, using large pretrained language models based on the Transformer architecture \cite{transformers}.
Namely, we finetune OpenAI-GPT \cite{OpenAI-GPT} and BERT \cite{BERT}, which have both shown remarkable improvements on a variety of tasks.
OpenAI-GPT is a uni-directional language model trained on the BookCorpus \cite{Zhu2015AligningBA}, whereas BERT is a bidirectional language model trained on both the BookCorpus and English Wikipedia.
As per previous work, we finetune the language model representations but fully learn the classifier specific parameters described below.

\paragraph{Multiple choice classification}
To classify sequences using these language models, we follow the multiple-choice setup implementation by the respective authors, as described below.
First, we concatenate the context, question, and answer, using the model specific separator tokens.
For OpenAI-GPT, the format becomes \texttt{\_start\_ <context>  <question> \_delimiter\_ <answer> \_classify\_}, where \texttt{\_start\_}, \texttt{\_delimiter\_}, and \texttt{\_classify\_} are special function tokens.
For BERT, the format is similar, but the classifier token comes before the context.\footnote{BERT's format is \texttt{[CLS] <context> [UNUSED] <question> [SEP] <answer> [SEP]}} 

For each triple, we then compute a score $l$ by passing the hidden representation from the classifier token $h_{CLS} \in \mathbb{R}^{H}$ through an MLP: 
$$l = W_2\tanh( W_1 h_{CLS} + b_1)$$
where $W_1 \in \mathbb{R}^{H\times H}$, $b_1 \in \mathbb{R}^H$ and $W_2 \in \mathbb{R}^{1\times H}$.
Finally, we normalize scores across all triples for a given context-question pair using a softmax layer.
The model's predicted answer corresponds to the triple with the highest probability.

\begin{table}[t]
    \centering
    \large
    \begin{tabular}{@{}lll@{}}
        \toprule
        \multirow{2}{*}{Model} & \multicolumn{2}{l@{}}{~Accuracy (\%)}\\
        & Dev & Test \\ \midrule
        Random baseline & 33.3 & 33.3 \\
        GPT &  63.3 & 63.0\\
        BERT-base & 63.3 & 63.1 \\
        BERT-large & {\bf 66.0} & {\bf 64.5} \\
        \midrule
        ~~~ w/o context & 52.7 & --\\
        ~~~ w/o question & 52.1 & --\\
        ~~~ w/o context, question & 45.5 & --\\
        \midrule
        Human & 86.9* & 84.4* \\ \bottomrule
    \end{tabular}
    \caption{Experimental results. We additionally perform an ablation by removing contexts and questions, verifying that both are necessary for BERT-large's performance.
    Human evaluation results are obtained using 900 randomly sampled examples.}    \label{tab:modelling-results}
\end{table}

\newcommand{\robot}{\includegraphics[height=.3cm]{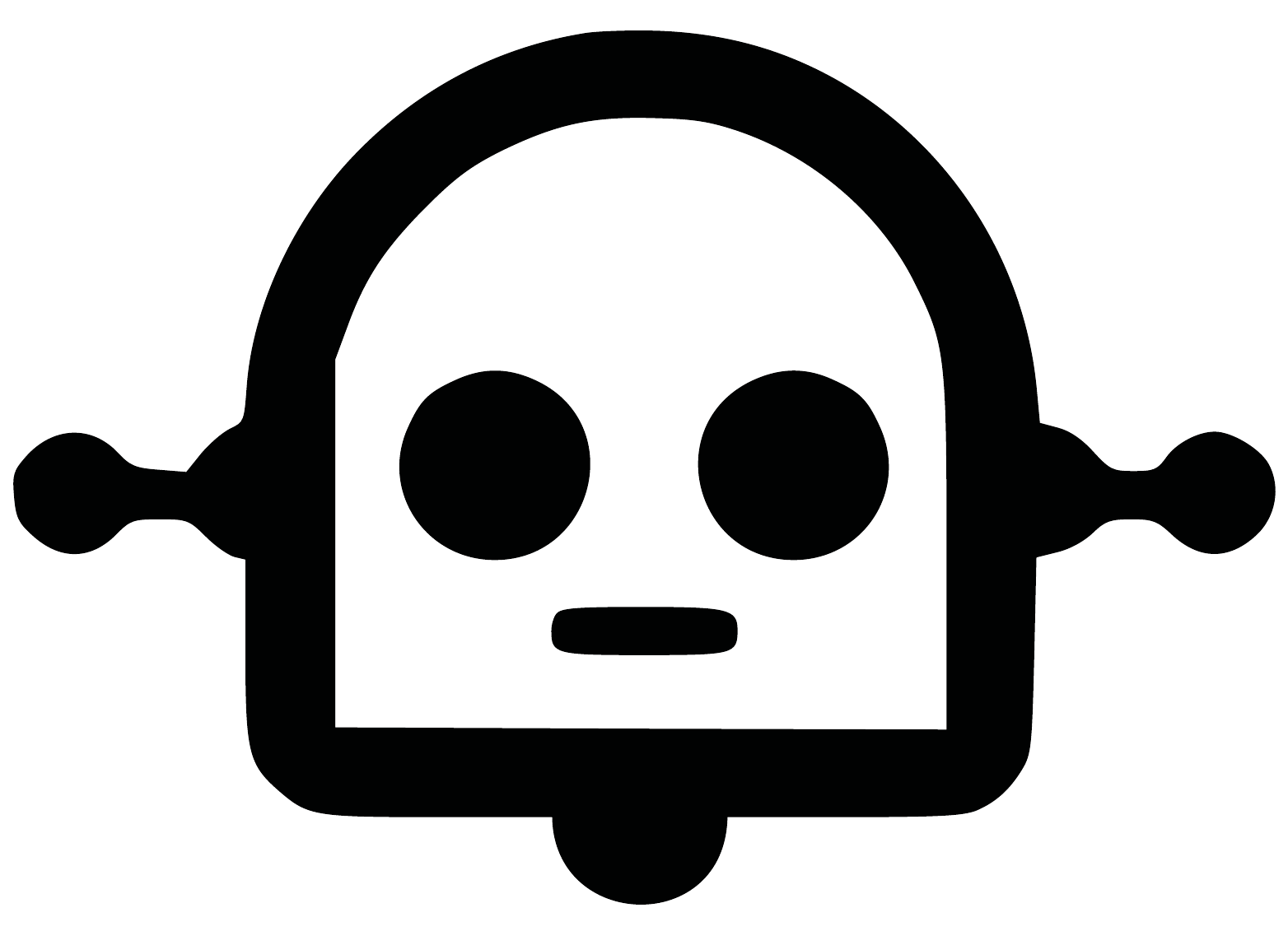}}
\newcommand{\qsasymbol}{}
\newcommand{\hiasymbol}{}

\begin{table*}[t]
\begin{center}
\newcommand{\contextColumnWidth}{5.15cm}
\newcommand{\questionColumnWidth}{2.6cm}
\newcommand{\answerColumnWidth}{5.8cm}

\begin{tabular}{@{}l@{\hspace{5px}}ll@{\hspace{5px}}c@{\hspace{3px}}c@{\hspace{5px}}p{\answerColumnWidth}@{}}
\toprule
 & Context & Question &  &  & Answer  \\ \midrule
 
\multirow{3}{*}{(1)} & \multirow{3}{\contextColumnWidth}{Jesse was pet sitting for Addison, so Jesse came to Addison's house and walked their dog.}&
\multirow{3}{\questionColumnWidth}{What does Jesse need to do before this?}&
&&(a) feed the dog\\
 &&&\checkmark&\robot&(b) get a key from Addison \\
 &&&&&(c) walk the dog\\\midrule

 
\multirow{3}{*}{(2)} & \multirow{3}{\contextColumnWidth}{Kai handed back the computer to Will after using it to buy a product off Amazon.}&
\multirow{3}{\questionColumnWidth}{What will Kai want to do next?}&
 &&(a) wanted to save money on shipping\\
 &&&\checkmark&\robot& (b) Wait for the package \\
 &&&&& (c) Wait for the computer \\\midrule
 
\multirow{3}{*}{(3)} & \multirow{3}{\contextColumnWidth}{Remy gave Skylar, the concierge, her account so that she could check into the hotel.}&
\multirow{3}{\questionColumnWidth}{What will Remy want to do next?}&&&
(a) lose her credit card\\
 &&&&\robot& (b) arrive at a hotel\\
 &&&\checkmark&&(c) get the key from Skylar\\\midrule

\multirow{3}{*}{(4)} & \multirow{3}{\contextColumnWidth}{Sydney woke up and was ready to start the day. They put on their clothes.}&
\multirow{3}{\questionColumnWidth}{What will Sydney want to do next?}&
 &\robot&(a) go to bed \\
 &&&&& (b) go to the pool \\
 &&&\checkmark&& (c) go to work \\\midrule
 

\multirow{3}{*}{(5)}&\multirow{3}{\contextColumnWidth}{Kai grabbed Carson's tools for him because Carson could not get them.}&\multirow{3}{\questionColumnWidth}{How would Carson feel as a result?}&&\robot&(a) inconvenienced\\
 & & &$\checkmark$& &  (b) grateful\\
 & & && & (c) angry \\\midrule
 
\multirow{3}{*}{(6)}& \multirow{3}{\contextColumnWidth}{Although Aubrey was older and stronger, they lost to Alex in arm wrestling.} & 
    \multirow{3}{\questionColumnWidth}{How would Alex feel as a result?} & && 
    (a) they need to practice more \qsasymbol\\
 &&&&\robot& (b) ashamed \hiasymbol  \\
 &&& $\checkmark$ & & (c) boastful  \\
 \bottomrule
\end{tabular}
\end{center}
\caption{\label{tab:data-examples} Example CQA triples from the \papername dev set with BERT-large's predictions
(\robot: BERT's prediction, $\checkmark$: true correct answer).  The model predicts correctly in (1) and (2) and incorrectly in the other four examples shown here.
Examples (3) and (4) illustrate the model choosing answers that might have happened before, or that might happen much later after the context, as opposed to right after the context situation. In Examples (5) and (6), the model chooses answers that may apply to people other than the ones being asked about. 
}
\end{table*}

\section{Experiments}
\subsection{Experimental Set-up}\label{ssec:experimental-setup}
We train our models on the 33k \papername training instances, selecting hyperparameters based on the best performing model on our dev set, for which we then report test results.
Specifically, we perform finetuning through a grid search over the hyper-parameter settings (with a learning rate in $\{1\mathrm{e}{-5}, 2\mathrm{e}{-5}, 3\mathrm{e}{-5}\}$, a batch size in $\{3, 4, 8\}$, and a number of epochs in $\{3, 4, 10\}$) and report the maximum performance. 

Models used in our experiments vary in sizes: OpenAI-GPT (117M parameters) has a hidden size $H$=768, 
BERT-base (110M params) and BERT-large (340M params) hidden sizes of $H$=768 and $H$=1024, respectively.
We train using the HuggingFace PyTorch \cite{paszke2017automatic} implementation.\footnote{\url{https://github.com/huggingface/pytorch-pretrained-BERT}}

\subsection{Results}
Our results (Table \ref{tab:modelling-results}) show that \papername is still a challenging benchmark for existing computational models, compared to human performance.
Our best performing model, BERT-large, outperforms other models by several points on the dev and test set. 
We additionally ablate our best model's representation by removing the context and question from the input, confirming that reasoning over both is necessary for this task.

\begin{figure}[t]
    \centering
    
    \includegraphics[width=\columnwidth]{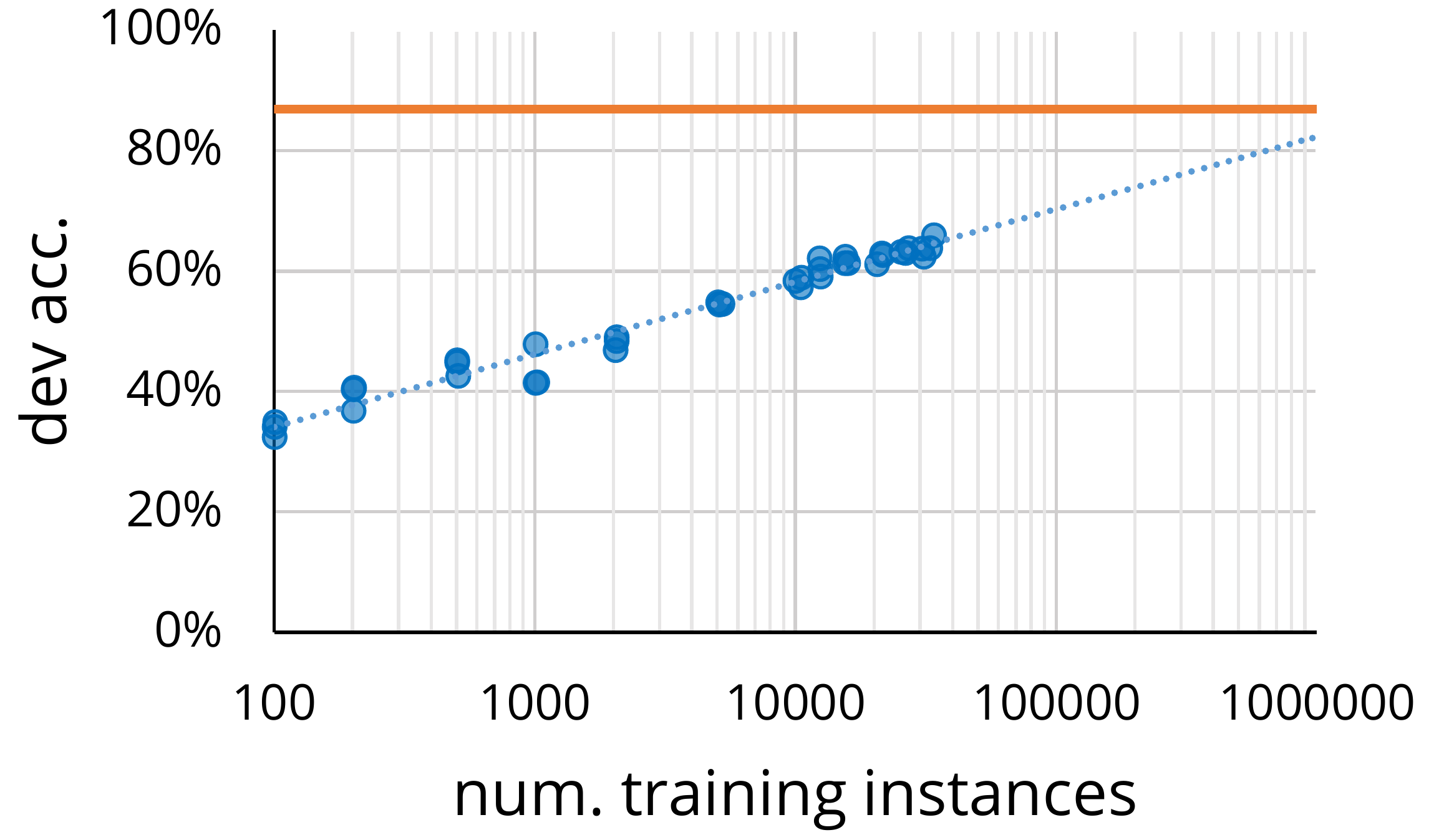}
    
    \caption{Dev accuracy when training BERT-large with various number of examples (multiple runs per training size), with human performance (86.9\%) shown in orange.
    In order to reach $>$80\%, the model would require nearly 1 million training examples.
    }
    \label{fig:learningcurve}
\end{figure}

\paragraph{Learning Curve}
To better understand the effect of dataset scale on model performance on our task, we simulate training situations with limited knowledge.
We present the learning curve of BERT-large's performance on the dev set as it is trained on more training set examples (Figure~\ref{fig:learningcurve}).
Although the model does significantly improve over a random baseline of 33\% with only a few hundred examples, the performance only starts to converge after around 20k examples, providing evidence that large-scale benchmarks are required for this type of reasoning.

\begin{figure}[t]
    \centering
    \includegraphics[width=.95\columnwidth]{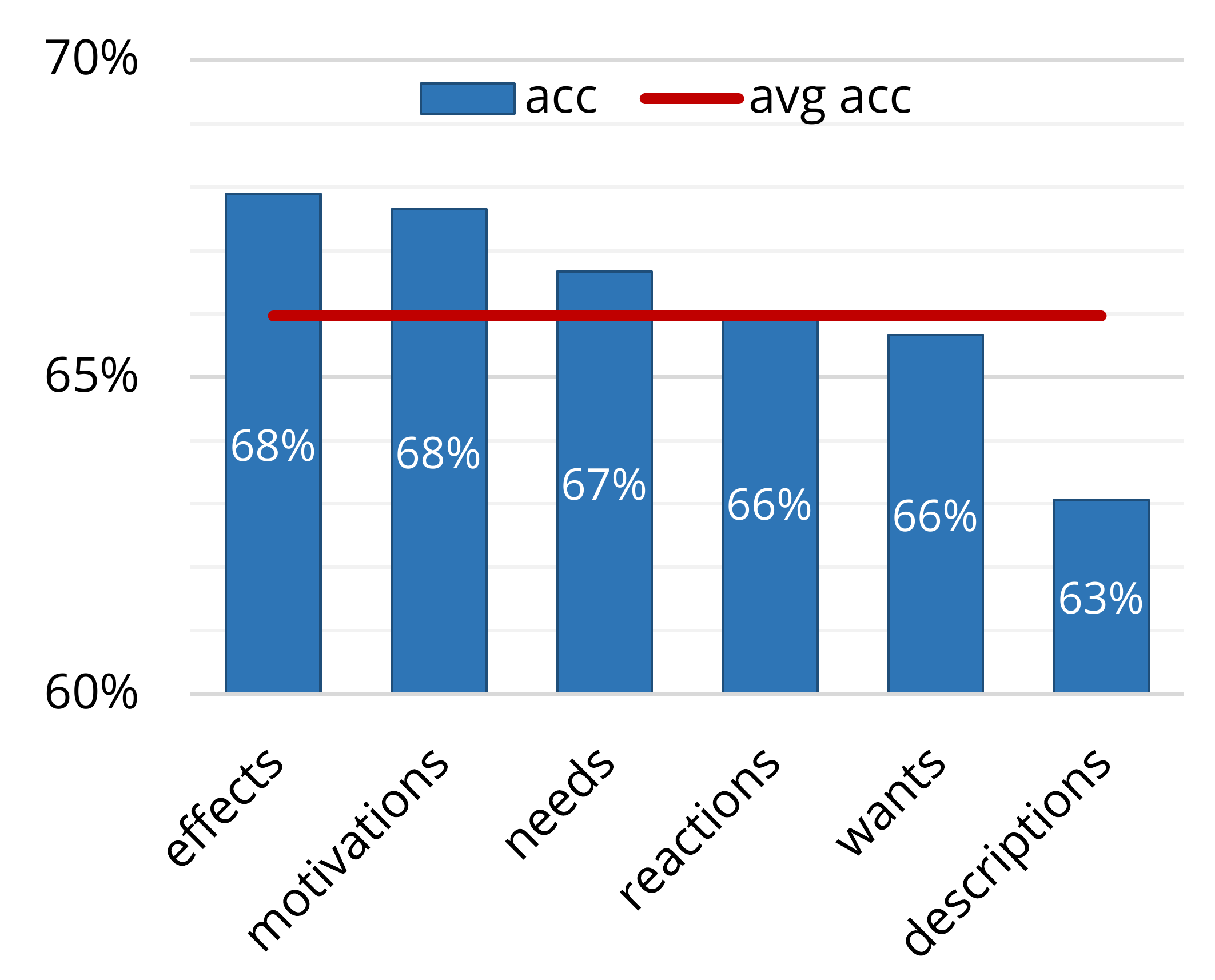}
    \caption{Average dev accuracy of BERT-large on different question types.
    While questions about effects and motivations are easier, the model still finds wants and descriptions more challenging.
    }
    \label{fig:bert-breakdown}
\end{figure}

\paragraph{Error Analysis} 
We include a breakdown of our best model's performance on various question types in Figure \ref{fig:bert-breakdown} and specific examples of errors in the last four rows of Table~\ref{tab:data-examples}.
Overall, questions related to pre-conditions of the context (people's motivations, actions needed before the context) are less challenging for the model.
Conversely, the model seems to struggle more with questions relating to (potentially involuntary) effects, stative descriptions, and what people will want to do next.

Examples of errors in Table~\ref{tab:data-examples} further indicate that, instead of doing advanced reasoning about situations, models may only be learning lexical associations between the context, question, and answers, as hinted at by \citet{Marcus2018DeepLA} and \citet{zellers2019hellaswag}.
This leads the model to select answers with incorrect timing (examples 3 and 4) or answers pertaining to the wrong participants (examples 5 and 6), despite being trained on large amounts of examples that specifically distinguish proper timing and participants.
For instance, in (3) and (4), the model selects answers which are incorrectly timed with respect to the context and question (e.g., ``arrive at a hotel'' is something Remy likely did before checking in with the concierge, not afterwards).
Additionally, the model often chooses answers related to a person other than the one asked about.
In (6), after the arm wrestling, though it is likely that Aubrey will feel ashamed, the question relates to what Alex might feel--not Aubrey.

Overall, our results illustrate how reasoning about social situations still remains a challenge for these models, compared to humans who can trivially reason about the causes and effects for multiple participants.
We expect that this task would benefit from models capable of more complex reasoning about entity state, 
or models that are more explicitly endowed with commonsense (e.g., from knowledge graphs like ATOMIC).



\section{\papername for Transfer Learning}
In addition to being the first large-scale benchmark for social commonsense, 
we also show that \papername can improve performance on downstream tasks that require commonsense, namely the Winograd Schema Challenge and the Choice of Plausible Alternatives task.
We achieve state of the art performance on both tasks by sequentially finetuning on \papername before the task itself.

\paragraph{COPA}
The Choice of Plausible Alternatives task \citep[COPA;][]{Roemmele2011ChoiceOP} is a two-way multiple choice task which aims to measure commonsense reasoning abilities of models.
The dataset contains 1,000 questions (500 dev, 500 test) that ask about the causes and effects of a premise.
This has been a challenging task for computational systems, partially due to the limited amount of training data available.
As done previously \citep{copacetic,luo2016commonsense}, we finetune our models on the dev set, and report performance only on the test set.

\paragraph{Winograd Schema}
The Winograd Schema Challenge \citep[WSC;][]{Levesque2011TheWS} is a well-known commonsense knowledge challenge framed as a coreference resolution task.
It contains a collection of $273$ short sentences in which a pronoun must be resolved to one of two antecedents 
(e.g., in ``The city councilmen refused the demonstrators a permit because \textit{they} feared violence'', \textit{they} refers to the councilmen).
Because of data scarcity in WSC,
\citet{Rahman2012DPR} created $943$ Winograd-style sentence pairs ($1886$ sentences in total), henceforth referred to as DPR, which has been shown to be slightly less challenging than WSC for computational models. 

We evaluate on these two benchmarks. While the DPR dataset is split into train and test sets \citep{Rahman2012DPR}, the WSC dataset contains a single (\emph{test}) set of only $273$ instances for evaluation purposes only.
Therefore, we use the DPR dataset as training set when evaluating on the WSC dataset.


\newcommand{\colspaceC}{\hspace{.3cm}}

\begin{table}[t]
    \centering
    \begin{tabular}{@{}c@{\colspaceC}ll@{\colspaceC}c@{\colspaceC}c@{}}
    \toprule
    \multirow{2}{*}{Task} & \multirow{2}{*}{Model} & \multicolumn{3}{@{}c@{}}{Acc. (\%)} \\
    & & best & mean & std \\
    \midrule
    \multirow{3}{*}{\rotatebox[origin=c]{90}{COPA}} & \citet{Sasaki2017HandlingME} & 71.2 & -- & --  \\
     & BERT-large & 80.8 & 75.0 & 3.0 \\
    & BERT-\papername &  \textbf{83.4} & 80.1 & 2.0\\
    \midrule
    \multirow{3}{*}{\rotatebox[origin=c]{90}{WSC}} & \citet{Kocijan2019ASR} & 72.5 & -- & -- \\ 
    & BERT-large & 67.0 & 65.5 & 1.0  \\
    & BERT-\papername & \textbf{72.5} & 69.6 & 1.7\\
    \midrule
    \multirow{3}{*}{\rotatebox[origin=c]{90}{DPR}}& \citet{Peng2015SolvingHC} & 76.4 & -- & --\\
    & BERT-large & 79.4  & 71.2 & 3.8 \\
    & BERT-\papername & \textbf{84.0} & 81.7 & 1.2\\
    \bottomrule
    \end{tabular}
    \caption{\label{tab:sequential-finetuning}Sequential finetuning of BERT-large on \papername before the task yields state of the art results (bolded) on COPA \citep{Roemmele2011ChoiceOP}, Winograd Schema Challenge \citep{Levesque2011TheWS} and DPR \citep{Rahman2012DPR}.
    For comparison, we include previous published state of the art performance.
    }
\end{table}

\subsection{Sequential Finetuning}
We first finetune BERT-large on \papername, which reaches 66\% on our dev set (Table~\ref{tab:modelling-results}).
We then finetune that model further on the task-specific datasets, considering the same set of hyperparameters as in \S \ref{ssec:experimental-setup}.
On each of the test sets, we report best, mean, and standard deviation of all models, and compare sequential finetuning results to a BERT-large baseline.




\paragraph{Results}
Shown in Table \ref{tab:sequential-finetuning}, sequential finetuning on \papername yields substantial improvements over the BERT-only baseline (between $2.6$ and $5.5\%$ max performance increases), as well as the general increase in performance stability (i.e., lower standard deviations).
As hinted at by \citet{Phang2019SentenceEO}, this suggests that BERT-large can benefit from both the large scale and the QA format of commonsense knowledge in \papername, which it struggles to learn from small benchmarks only.
Notably, we find that sequentially finetuned BERT-\papername achieves state-of-the-art results on all three tasks, showing improvements of previous best performing models.\footnote{
Note that OpenAI-GPT was reported to achieve 78.6\% on COPA, but that result was not published, nor discussed in the OpenAI-GPT white paper \cite{OpenAI-GPT}.}

\begin{figure}
    \centering
    \includegraphics[width=\columnwidth]{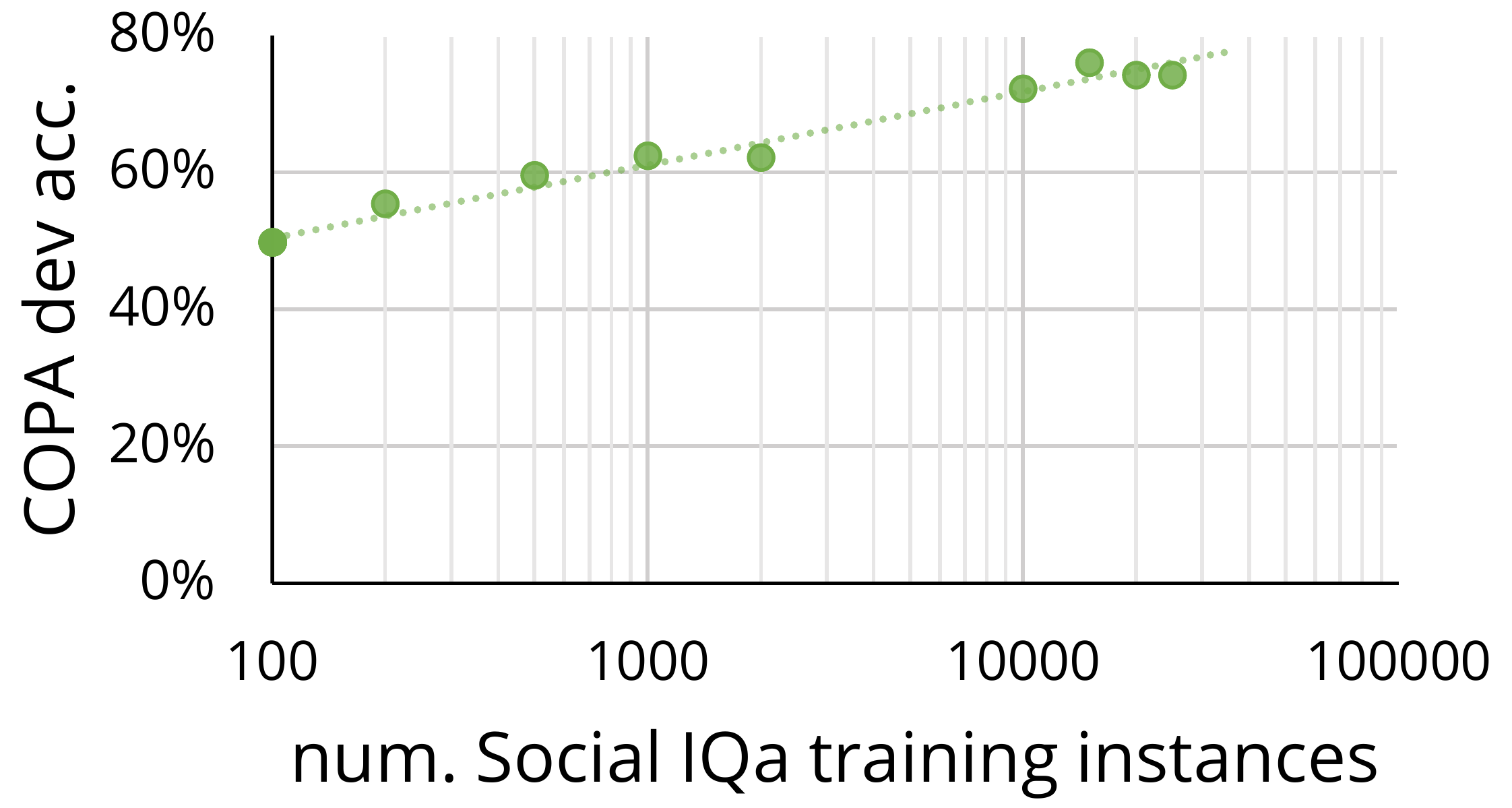}
    \caption{Effect of finetuning BERT-large on varying sizes of the \papername training set on the dev accuracy of COPA.
    As expected, the more \papername instances the model is finetuned on, the better the accuracy on COPA.}
    \label{fig:siqa_sizes_copa}
\end{figure}

\paragraph{Effect of scale and knowledge type}
To better understand these improvements in downstream task performance, we investigate the impact on COPA performance of sequential finetuning on less \papername training data (Figure~\ref{fig:siqa_sizes_copa}), as well as the impact of the type of commonsense knowledge used in sequential finetuning.
As expected, the downstream performance on COPA improves when using a model pretrained on more of \papername, indicating that the scale of the dataset is one factor that helps in the fine-tuning.
However, when using SWAG (a similarly sized dataset) instead of \papername for sequential finetuning, the downstream performance on COPA is lower (76.2\%).
This indicates that, in addition to its large scale, the social and emotional nature of the knowledge in \papername enables improvements on these downstream tasks.

\section{Related Work}
\paragraph{Commonsense Benchmarks:} 
Commonsense benchmark creation has been well-studied by previous work. 
Notably, the WinoGrad Schema Challenge \cite[WSC;][]{Levesque2011TheWS} and the Choice Of Plausible Alternatives dataset  \cite[COPA;][]{Roemmele2011ChoiceOP} are expert-curated collections of commonsense QA pairs that are trivial for humans to solve.
Whereas WSC requires physical and social commonsense knowledge to solve, COPA targets the knowledge of causes and effects surrounding social situations.
While both benchmarks are of high-quality and created by experts, their small scale (150 and 1,000 examples, respectively) poses a challenge for modern modelling techniques, which require many training instances.

More recently, \citet{Talmor2019commonsenseQA} introduce CommonsenseQA, containing 12k multiple-choice questions.
Crowdsourced using ConceptNet \cite{ConceptNet}, these questions mostly probe knowledge related to factual and physical commonsense (e.g., ``Where would I not want a fox?'').
In contrast, \papername explicitly separates contexts from questions, and focuses on the types of commonsense inferences humans perform when navigating social situations.

\paragraph{Commonsense Knowledge Bases:}
In addition to large-scale benchmarks, there is a wealth of work aimed at creating commonsense knowledge repositories \citep{ConceptNet,sap2019atomic,zhang2017ordinal,lenat1995cyc,eventNet,gordon2017formal} that can be used as resources in downstream reasoning tasks. 
While \papername is formatted as a natural language QA benchmark, rather than a taxonomic knowledge base, it also can be used as a resource for external tasks, as we have demonstrated experimentally.

\paragraph{Constrained or Adversarial Data Collection:} Various work has investigated ways to circumvent annotation artifacts that result from crowdsourcing.
\citet{Sharma2018TacklingTS} extend the Story Cloze data by severely restricting the incorrect story ending generation task, reducing the sentiment and negation artifacts.
\citet{rajpurkar2018know} create an adversarial version of the extractive question-answering challenge, SQuAD \cite{Rajpurkar2016SQuAD10}, by creating 50k unanswerable questions.
Instead of using human-generated incorrect answers, \citet{Zellers2018SWAGAL,zellers2019hellaswag} use adversarial filtering of machine generated incorrect answers to minimize surface patterns. Our dataset also aims to reduce annotation artifacts by using a multi-stage annotation pipeline in which we collect negative responses from multiple methods including a unique adversarial question-switching technique.


\section{Conclusion}
We present \papername, the first large-scale benchmark for social commonsense.
Consisting of 38k multiple-choice questions, \papername covers various types of inference about people's actions being described in situational contexts.
We design a crowdsourcing framework for collecting QA pairs that reduces stylistic artifacts of negative answers through an adversarial question-switching method.
Despite human performance of close to 90\%,
computational approaches based on large pretrained language models only achieve accuracies up to 65\%, suggesting that these social inferences are still a challenge for AI systems.
In addition to providing a new benchmark, we demonstrate how transfer learning from \papername to other commonsense challenges can yield significant improvements, achieving
new state-of-the-art performance on both COPA and Winograd Schema Challenge datasets.

\section*{Acknowledgments}
We thank Chandra Bhagavatula, Hannaneh Hajishirzi, and other members of the UW NLP and AI2 community for helpful discussions and feedback throughout this project. 
We also thank the anonymous reviewers for their insightful comments and suggestions.  This research was supported in part by NSF (IIS-1524371, IIS-1714566), DARPA under the CwC program through the ARO (W911NF-15-1-0543), DARPA under the MCS program through NIWC Pacific (N66001-19-2-4031), Samsung Research, and the National Science Foundation Graduate Research Fellowship Program under Grant No. DGE-1256082.

\bibliography{socialIQa}
\bibliographystyle{acl_natbib}


\end{document}